\documentclass[runningheads]{llncs}
\usepackage{graphicx}
\usepackage [autostyle, english = american]{csquotes}
\MakeOuterQuote{"}
\usepackage{algpseudocode,algorithm,algorithmicx}  

\begin{document}
\title{Deep Neural Network Ensembles}
%
%
\author{Sean Tao\inst{1}\orcidID{0000-0002-6765-4514}}

\institute{Carnegie Mellon University, Pittsburgh PA 15213, USA
\email{shtao@alumni.cmu.edu}\\}
\maketitle              
\begin{abstract}
Current deep neural networks suffer from two problems; first, they are hard to interpret, and second, they suffer from overfitting. There have been many attempts to define interpretability in neural networks, but they typically lack causality or generality. A myriad of regularization techniques have been developed to prevent overfitting, and this has driven deep learning to become the hot topic it is today; however, while most regularization techniques are justified empirically and even intuitively, there is not much underlying theory. This paper argues that to extract the features used in neural networks to make decisions, it's important to look at the paths between clusters existing in the hidden spaces of neural networks. These features are of particular interest because they reflect the true decision-making process of the neural network. This analysis is then furthered to present an ensemble algorithm for arbitrary neural networks which has guarantees for test accuracy. Finally, a discussion detailing the aforementioned guarantees is introduced and the implications to neural networks, including an intuitive explanation for all current regularization methods, are presented. The ensemble algorithm has generated state-of-the-art results for Wide-ResNets on CIFAR-10 (top 5 for all models) and has improved test accuracy for all models it has been applied to.

\end{abstract}

\section{Introduction}
 Consider a simple feed forward neural network. Define a hidden space corresponding to a hidden layer of a neural network as the space containing the outputs of the hidden nodes at that hidden layer for some input. All hidden spaces are composed of perceptrons with respect to the previous layer, each of which has a hyperplane decision boundary. Points in the previous space are mapped to a constant function of their distance to this plane, and depending on the activation function, compressed or stretched. This stretching and compressing naturally leads to clustering in hidden spaces. The process is repeated for each perceptron comprising the hidden space, where adding a perceptron adds a dimension to the hidden space by projecting the points into a new dimension depending on their distances from the hyperplane of the new perceptron and the activation function. Define a feature to be a measurable characteristic which a neural network uses to make its classification decision. Unfortunately, these clusters are not features in and of themselves but rather mixtures of features. To extract individual features, then, cluster paths should be examined, where a path is defined per individual input point as the sequence of clusters in the neural network the point belongs to, starting from cluster it belongs to in the input space, then the clusters it belongs to in each hidden space, and finally the cluster it belongs to in the output space.

 Intuitively, paths represent features because each path defines a region of the input space which will eventually end up at the same cluster in the output space via the same logic pathway used by the neural network. Thus, it immediately follows that a point on a path must be classified similarly as other points on that path, assuming all points in the output space are classified by cluster. This is another way of stating that points on the same path are indistinguishable from each other to the neural network, and since points in different paths were differentiated in some layer, paths separate out features of neural networks. This process is formalized in the algorithm described below. However, besides merely finding the features in a neural network, the paths serve an even more important purpose--they separate the input into regions of confidence with respect to the output classification. In particular, paths represent specific features, some of which were found because they are truly useful and others due to overfitting.
 
 The process of determining "good" and "bad" data points, formally defined below, attempts to separate data into these two categories. Informally, "good" data come from paths that contain many points that are classified correctly, since these are likely not due to random chance and thus are real features. An ensemble algorithm can then be created to combine different models, where models only vote on their "good" data points. This is equivalent to querying neural networks for points where they are confident in their predictions. Points where the model is unsure can then be classified by other models.

\section{Related Work}
Much effort has been put into interpreting deep neural networks in the past \cite{olah2017feature}. A few meta studies summarize the effort of the community as a whole rather well. There are three general types of methods for deep neural networks; namely, by discovering "ways to reduce the complexity of all these operations," "understand[ing] the role and structure of the data flowing through these bottlenecks," and "creat[ing] networks that are designed to be easier to explain" \cite{Gilpin2018ExplainingEA}. The difference between the algorithm presented here and the aforementioned attempts at interpretability is that this derives its logic entirely from the network--in particular, paths, by definition, represent the features used in classification.

In addition, clustering has been applied to the hidden spaces of neural networks in prior work, also in the context of interpretability \cite{Liu2018InterpretableDC}. However, cluster path analysis in deep neural networks could not be found.

Finally, ensembling algorithms of deep neural networks have been attempted before. For instance, algorithms already presented in different contexts have been applied to neural networks \cite{Ju2017TheRP}, such as the Super Learner  \cite{Laan2007SuperL} and AdaBoost \cite{Schwenk2000BoostingNN}. Nevertheless, "the behavior of ensemble learning with deep networks is still not well studied and understood" \cite{Ju2017TheRP}.

\section{Algorithm}

\subsection{Training}

Here, the algorithm to train the ensemble is presented; this also contains the process to generate paths. As clarification, the distinction in this paper between training data and a training set is that the training data is split into a training set and a validation set. First, partition the training data into training/validation sets. Repeat this such that there is minimal overlap between points in different training sets. Then, for each of the partitions, conduct the following steps.

Train a neural network on the training set, selecting the best model via validation accuracy. Add this neural network to the set of models named model 1’s ("original models"). Determine the optimal number of means in each layer via the "elbow" technique \cite{Thorndike1953} by plotting inertia against the number of means with k-means \cite{zbMATH03129892}. Run k-means again with the optimal number of means on the input layer, each of the hidden layers, and output layer for the training set. For each point in the training and validation sets, determine the path of clusters through the neural network. Find optimal values of the following three parameters via grid search, filter out validation points which do not meet the criteria, and calculate the validation accuracy on the remaining points. The three parameters are: maximum distance to a cluster center, minimum number of data points in a split, and minimum accuracy in a split. Define a split to be a partial path of length 2--in other words, how one cluster was split into different clusters in the subsequent layer. Thus, for each point in the validation set, if it is too far from its cluster in any layer, if it is in a split which has too few points, or if it is in a split with too low validation accuracy, filter out the point, and calculate the validation accuracy on the remaining points. Ideally, the parameters which represent the tightest restrictions that filter out the fewest points and achieve the desired validation accuracy should be chosen. This process identifies paths which represent features not generated by overfitting and contain points of high validation, and thus test, accuracy. The idea is to then train other models to focus specifically on the "bad" test points. Using the same parameters found above, separate the training set into "good" and "bad" training points, where "good" data points are points which satisfy the parameters found above. Train another neural network, with the same partition of train and validation sets. However, repeat each "bad" training point in the training set, such that each "bad" point appears twice. Add this network to the set of models named model 2’s ("bad 1 models"). Repeat the analysis steps on this new neural network.

\begin{algorithm}  
  \caption{Training the Ensemble}  
  \begin{algorithmic}[5]  
    \State Partition the training data into training/validation sets
    \For{each partition}  
        \State Train a neural network
        \State Add this network to the set of model 1’s ("original models")
        \State Run k-means on all layers
        \For{each point in training and validation sets}
            \State Determine the path of clusters
        \EndFor
        \State Find parameters for filtering clusters
        \State Separate the training set into "good" and "bad" points
        \State Train another neural network with oversampling on the "bad" points
        \State Add this network to the set of model 2’s ("bad 1 models")
        \State Repeat the analysis steps for the new neural network
    \EndFor
  \end{algorithmic}  
\end{algorithm}

\subsection{Testing}
To run this ensemble on the test set, for each test data point, conduct the following steps. Determine whether or not it’s a "good" data point in each of the models in the set of model 1’s by using the parameters for the respective model. If it’s a "good" data point in at least half of the models in the set of model 1’s, and any of these models agree on a label, return that label. Call these test points "original good" test points. Otherwise, determine whether or not it’s a "good" test point in each of the models in the set of model 2’s. If it’s a "good" test point in at least half of the models in the set of model 2’s, and if any of these models agree on a label, return that label. Call these test points "bad 1" test points. Otherwise, add the output vectors of each of the models in the set of model 2’s, and return the label corresponding to the largest value in the resulting sum vector. Call these data points "bad 2" data points.

\begin{algorithm}  
  \caption{Testing the Ensemble}  
  \begin{algorithmic}[5]  
    \For{each test data point}  
        \If{"good" in set of model 1's and agree on label}  
          \State Classify via voting of "good" model 1's ("good" test point)
          \State \textbf{continue}
        \EndIf
        \If{"good" in set of model 2's and agree on label}  
          \State Classify via voting of "good" model 2's ("bad 1" test point)
          \State \textbf{continue}
        \EndIf
        \State Classify via majority voting of all model 2's ("bad 2" test point)
    \EndFor
  \end{algorithmic}  
\end{algorithm}

\subsection{Larger Models}
While this algorithm works to increase test accuracy in smaller, feed forward neural networks, it must be modified to so that it is effective for larger, state-of-the-art models. This is necessary for a few reasons. First, the architecture may not be able to be represented as a simple directed acyclic graph, and thus hidden spaces may not be defined. Second, even if some modern architectures avoid these problems, they often contain such high numbers of hidden nodes that clustering is not meaningful due to the curse of dimensionality. Fortunately, the concept of "good" and "bad" points seem to transcend model architectures and training processes. That is, models all seem to classify "good" points with much higher accuracy than "bad" points, regardless of their architecture, methods of regularization applied, or other training techniques. In particular, larger models seem to be able to accurately classify points which smaller models can classify correctly when trained on any subset of the training data. Then it follows that only "bad 2" test data are inaccurate for larger models. Thus, for larger models, to create an ensemble, conduct the following steps.

First, train one neural network normally ("original model"). This will be used to classify "good" and "bad 1" test data. Then, train another neural network ("bad model"), oversampling all training points which were classified as "bad" in the majority of the set of model 1's, as defined in the training section. Finally, use this model to classify all "bad 2" test data.

\begin{algorithm}  
  \caption{Training and Testing on Larger Models}  
  \begin{algorithmic}[5]
  \State Train larger model normally ("original model")
  \State Train larger model with oversampled "bad" training points as found in smaller models ("bad model")
    \For{each test data point}  
        \If{"good" or "bad 1" test point}  
          \State Classify using "original model"
          \State \textbf{continue}
        \EndIf
        \If{"bad 2" test point}
          \State Classify using "bad model"
        \EndIf
    \EndFor
  \end{algorithmic}  
\end{algorithm}

\section{Results}
The path analysis portion of the algorithm was run on a feed forward neural network with Dropout \cite{Srivastava2014DropoutAS} before each of 4 hidden layers and the input layer, all of which utilized sigmoid activations, trained with the Adam optimizer \cite{Kingma2015AdamAM} on the MNIST digits dataset \cite{lecun-mnisthandwrittendigit-2010}. Features were generated from all "good" splits in two manners: first, by averaging all the training points in that split; and second, by finding via backpropagation the input which best generates the cluster center at that layer while the hidden layers' weights remain fixed. No additional regularization is used, demonstrating an improvement over existing techniques, since typically this would just generate noise \cite{olah2017feature}. The former technique was used on the splits from the first to second hidden layer, since the previous splits had too many images to display, and subsequent layers only had 10 clusters, one per label. Due to lack of resources, cluster split means could not be found, but those could potentially generate clearer features. The latter technique was used on splits from the input to the first hidden layer. For both methods, visual inspection confirms that the features generated separate different digits and, when they exist, different manners of writing each digit. The features are presented below.
\begin{figure*}[ht!]
            \includegraphics[width=40px]{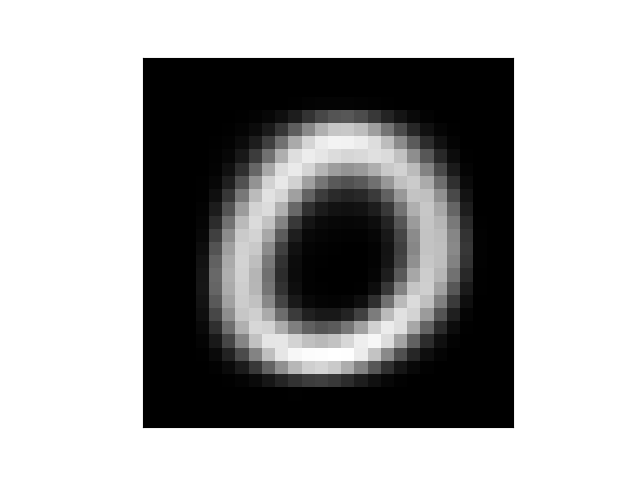}\hfill
            \includegraphics[width=40px]{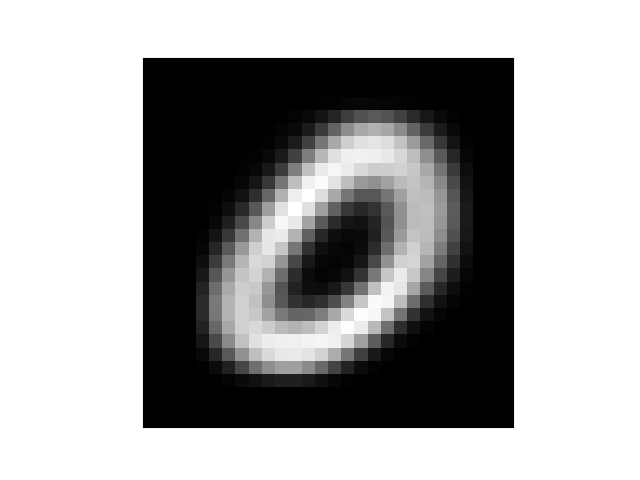}\hfill
            \includegraphics[width=40px]{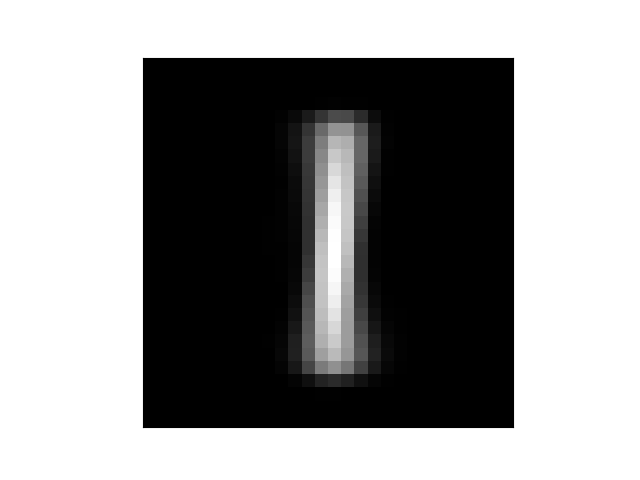}\hfill
            \includegraphics[width=40px]{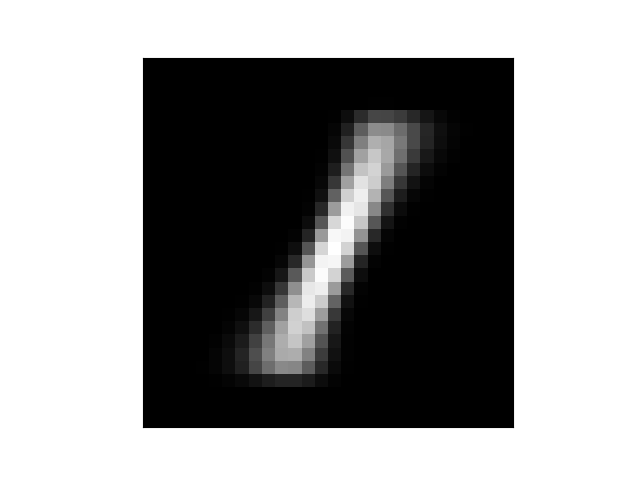}\hfill
            \includegraphics[width=40px]{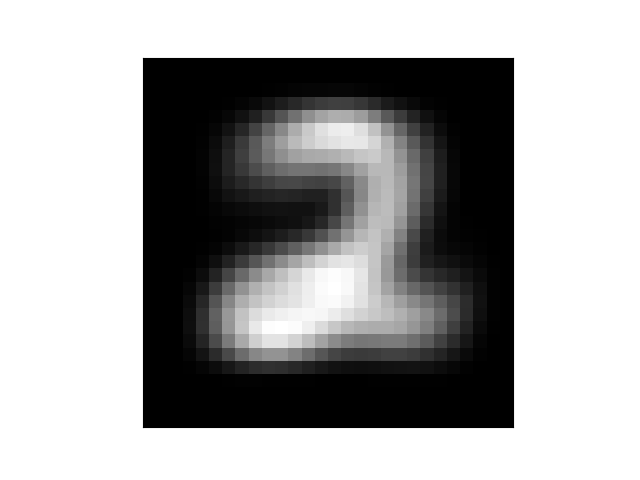}\hfill
            \includegraphics[width=40px]{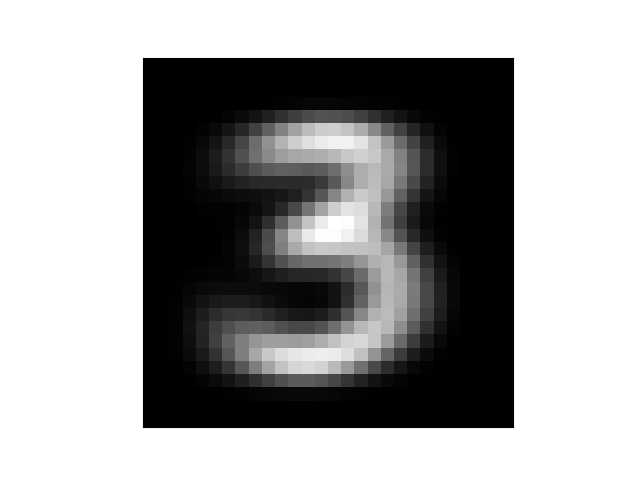}\hfill
            \includegraphics[width=40px]{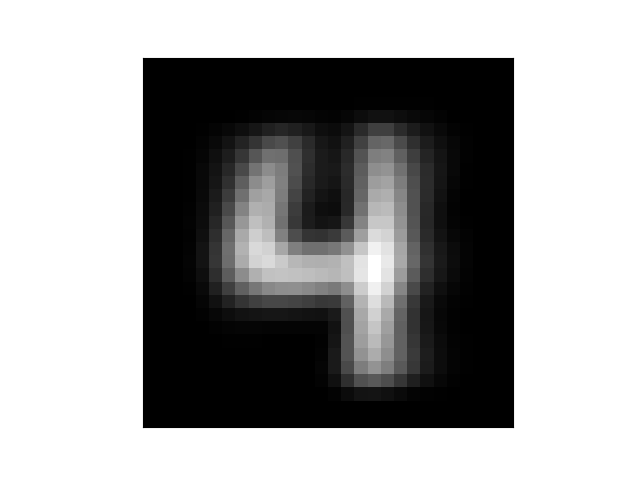}\hfill
            \includegraphics[width=40px]{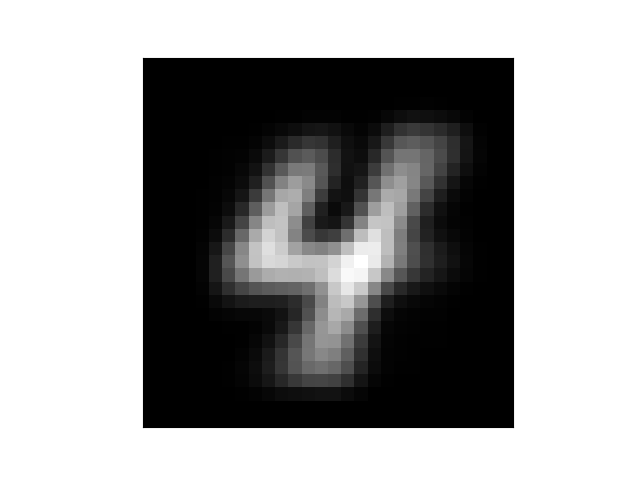}\hfill
            \hspace{0px}
            
            \includegraphics[width=40px]{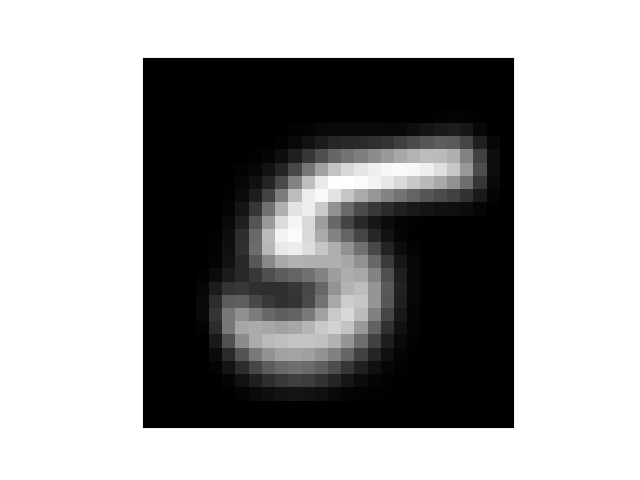}\hfill
            \includegraphics[width=40px]{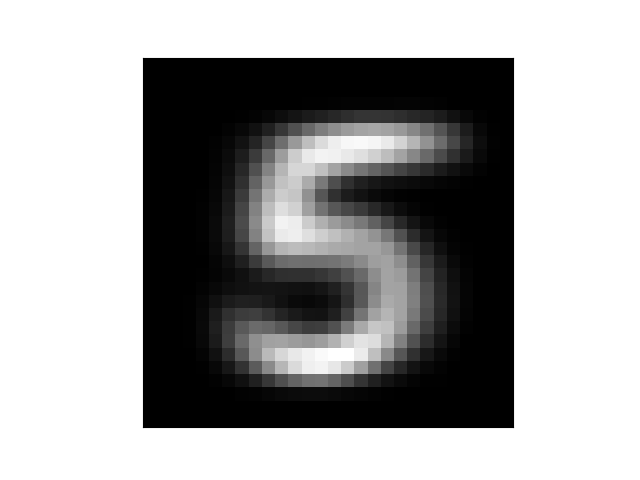}\hfill
            \includegraphics[width=40px]{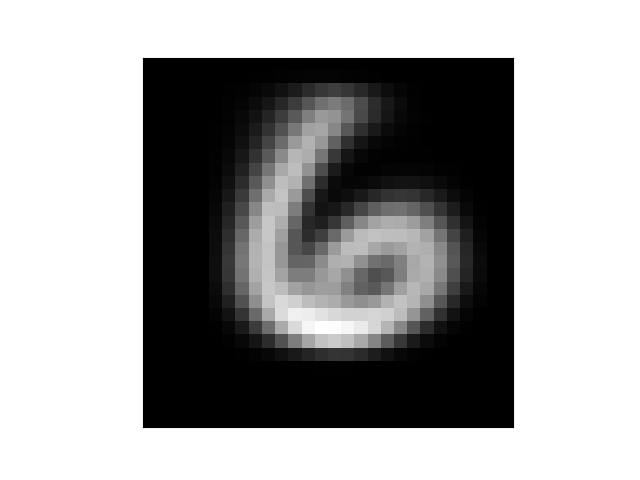}\hfill
            \includegraphics[width=40px]{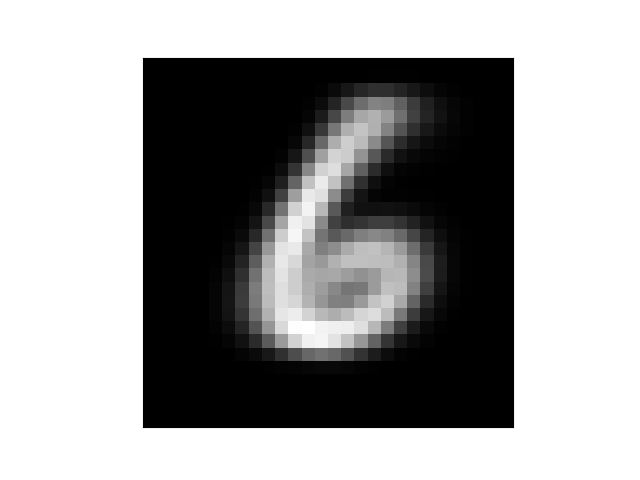}\hfill
            \includegraphics[width=40px]{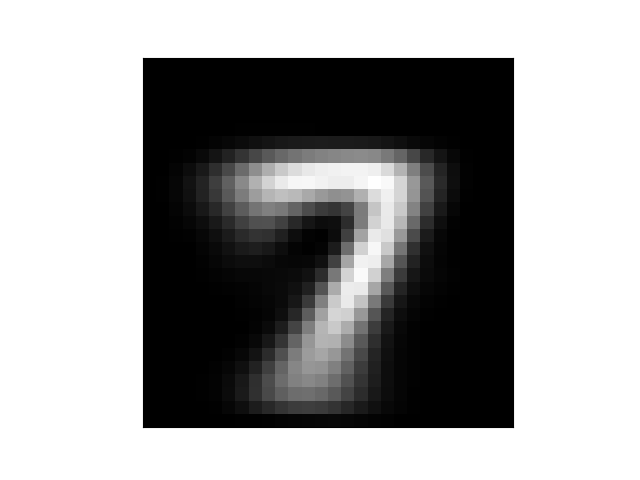}\hfill
            \includegraphics[width=40px]{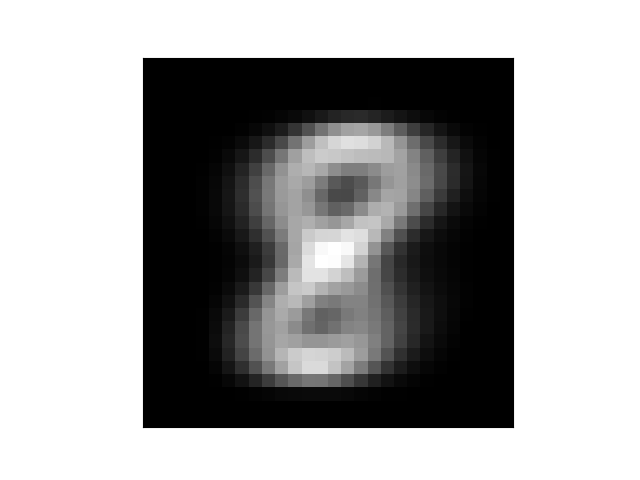}\hfill
            \includegraphics[width=40px]{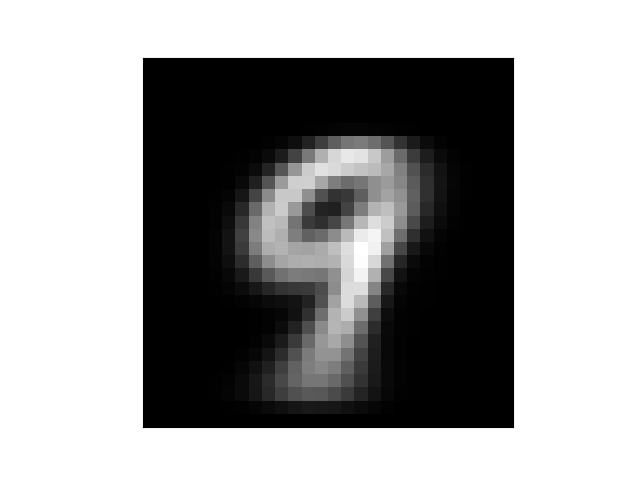}\hfill
            \includegraphics[width=40px]{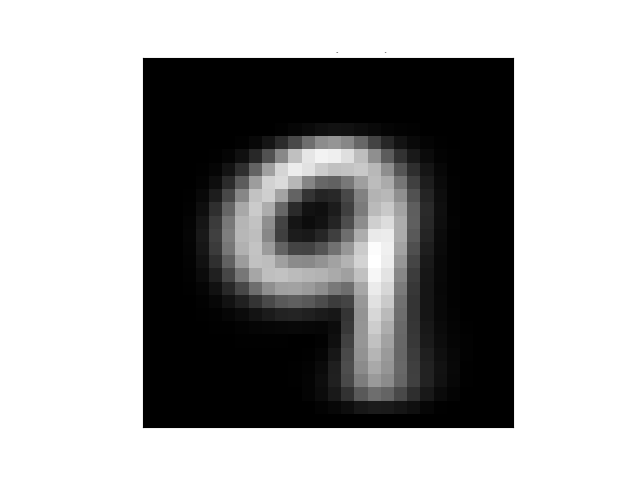}\hfill
            \caption{All "good" features from the first to the second hidden layer via input averaging}
        \end{figure*}
\begin{figure*}[ht!]
\centering
            \includegraphics[width=200px]{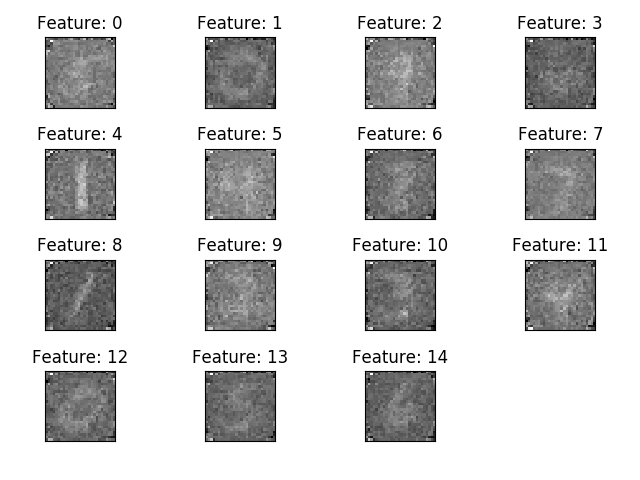}
            \caption{All "good" features from the input to the first hidden layer via backpropagation}
        \end{figure*}
The entire training algorithm was run on CIFAR-10 \cite{Krizhevsky2009LearningML} with an ensemble where the set of model 1's consisted of neural networks created by transfer learning of InceptionResNet \cite{Szegedy2016Inceptionv4IA} trained on ImageNet \cite{imagenet_cvpr09} and the set of model 2's consisted of VGG \cite{Simonyan2015VeryDC} models specific to CIFAR-10 \cite{cifar-vgg}. The highest test accuracy of any individual of these neural networks when trained on CIFAR-10 is 93.56\% \cite{cifar-vgg}. With five validation sets ("Block Partitions"), where each validation set was created by partitioning the training data into contiguous blocks of size 10000, the ensemble achieves 94.63\% test accuracy. With a different partition ("Stride 1 Partitions"), where the five validation sets were created by equivalence classes of the training data enumerated and taken modulo 5, the ensemble achieves 94.26\% test accuracy. When these two are combined and 10 validation sets are used, the accuracy increases to 94.77\%, a 1.21\% increase over any the test accuracy of individual network trained on the entire training data \cite{cifar-vgg}. The specific breakdown over the different sets of points are presented below. The difference between the test accuracy of "original good" points, which were "good" on the set of model 1's,  and "bad 1" points, which were "good" on the set of model 2's, and "bad 2" points is apparent.

\begin{center}
 \begin{tabular}{||c c c c c||} 
 \hline
 \multicolumn{5}{|c|}{Small Model Test Accuracy Breakdown} \\
 \hline
 Ensemble (Total Num Models) & "Original Good" & "Bad 1" & "Bad 2" & Overall \\ [0.5ex] 
 \hline
 \hline
  Block Partitions (10) & 98.97\% & 99.37\% & 79.65\% & 94.63\% \\ 
  \hline
 Stride 1 Partitions (10) & 99.11\% & 99.33\% & 76.71\% & 94.26\% \\ 
   \hline
 All Partitions (20) & 99.06\% & 99.29\% & 77.80\% & 94.77\% \\ 
 \hline
\end{tabular}
\end{center}

\begin{center}
 \begin{tabular}{||c c c c c||} 
 \hline
 \multicolumn{5}{|c|}{Small Model Total Number of Points by Type Breakdown  } \\
 \hline
 Ensemble (Total Num Models) & "Original Good" & "Bad 1" & "Bad 2" & Overall \\ [0.5ex] 
 \hline
 \hline
  Block Partitions (10) & 5815 & 1900 & 2285 & 10000 \\ 
  \hline
 Stride 1 Partitions (10) & 5255 & 2555 & 2190 & 10000 \\ 
   \hline
 All Partitions (20) & 5853 & 2106 & 2041 & 10000 \\ 
 \hline
\end{tabular}
\end{center}
Finally, the large model algorithm (algorithm 3) was applied a Wide-Res Net (WRN) \cite{Zagoruyko2016WideRN} trained with a data augmentation technique, AutoAugment \cite{Cubuk2018AutoAugmentLA}. This ensemble achieved 97.51\% test accuracy, 0.18\% better than the results for WRN's reported in the AutoAugment paper as of 9 October 2018 and a top 5 result of all papers on CIFAR-10. ShakeDrop \cite{Yamada2018ShakeDropRF} with AutoAugment achieved state-of-the-art results on CIFAR-10 when it was published, but the large model algorithm could not be applied to this due to resource limitations.
\begin{center}
 \begin{tabular}{||c c c c c||} 
 \hline
 \multicolumn{5}{|c|}{Large Model Ensemble Test Accuracy Breakdown} \\
 \hline
 Ensemble (Total Num Models) & "Original Good" & "Bad 1" & "Bad 2" & Overall \\ [0.5ex] 
 \hline
 \hline
  WRN (2) & 99.62\% & 99.57\% & 89.31\% & 97.51\% \\ 
  \hline
\end{tabular}
\end{center}

\begin{center}
 \begin{tabular}{||c c c c c||} 
 \hline
 \multicolumn{5}{|c|}{Large Model Ensemble Total Number of Points by Type Breakdown  } \\
 \hline
 Ensemble (Total Num Models) & "Original Good" & "Bad 1" & "Bad 2" & Overall \\ [0.5ex] 
 \hline
 \hline
  WRN (2) & 5853 & 2106 & 2041 & 10000 \\ 
 \hline
\end{tabular}
\end{center}

\section{Analysis}

\subsection{"Good" Paths}
\subsubsection{Bounds on Test Error}
Assume that the training data and the test data are independent and identically distributed, and the training data is sufficiently large. For neural networks, there exist potentially disjoint subspaces of the input space which correspond to "good" points in a path--namely, if a point were to fall into a particular subspace, then the point will be "good" in that is does not get filtered, and it will follow the corresponding path. Consider $k$ neural networks each trained on disjoint and insignificant fractions $f$ of the training data, and assume that each discovers some similar subspace. Due to the assumption that the training data is large, each of the $k$ neural networks is effectively trained on some subset of the data sampled from the distribution of all (train and test) data. Moreover, because the samples are disjoint, there is no dependence introduced by overlapping data.

Now, consider the intersection of all $k$ previously discussed subspaces, one found per network; call this s. Assume $n$ points from the respective validation set of each network reside in s, and the maximum classification error of the respective $n$ points for each of the $k$ networks is $\epsilon'$. Define $p$ to be the probability that a training set of size $f$ will produce a neural network that will classify all points in this subspace from the training and test points with error at most $\epsilon'$; in other words, the probability that a neural network will discover s. Noting that the maximum variance for a binary variable is $\frac{1}{4}$ and applying the central limit theorem, a confidence interval can be created to estimate $p$, since a sample mean is approximately normally distributed with variance at most $\frac{1}{4\sqrt{k}}$, and there exists a single sample mean with value 1. Define $\epsilon$ to be the average true error for all points in this subspace of any model trained on the fraction $f$ of the training data. Due to the properties of the normal distribution, the true value of $p$ is almost certainly within 6 standard deviations of this, or $1-\frac{6}{4\sqrt{k}}$. Then with high confidence, and by applying similar logic as above, the probability $\epsilon'$ is approximately normally distributed with mean $\epsilon$ and variance at most $\frac{1}{4\sqrt{n}}$ is $(1-\frac{6}{4\sqrt{k}})$, which converges to 1 as $k$ grows large. More concisely, for any confidence level, by using a sufficiently large ensemble, $\epsilon'$ is normally distributed with mean $\epsilon$ and variance at most $\frac{1}{4\sqrt{n}}$. This is important because now a confidence interval can be created for $\epsilon$, the average actual error of all points, train and test, in this subspace.

\begin{theorem}
Define $\epsilon$ to be the average true error for all points in some subspace of any model trained on the fraction $f$ of the training data. Consider $k$ neural networks each trained on disjoint and insignificant fractions $f$ of the training data, and assume that each discovers some similar subspace. Assume $n$ points from the respective validation set of each network reside in s, and the maximum classification error of the respective $n$ points for each of the $k$ networks is $\epsilon'$. Then with high confidence, the probability $\epsilon'$ is approximately normally distributed with mean $\epsilon$ and variance at most $\frac{1}{4\sqrt{n}}$ is $(1-\frac{6}{4\sqrt{k}})$, which converges to 1 as $k$ grows large.
\end{theorem}

\subsubsection{Implications}
The only previous work comparable to this requires either a finite hypothesis space or restrictions on the capabilities of the model--specifically the Vapnik-Chervonenkis (VC) dimension--and this is inapplicable to modern neural networks \cite{Vapnik2000TheNO}. This lies in between--it provides bounds on test error for extremely complicated models, but only for certain data points. The intuitive explanation behind the previous section is thus: a neural network trained on training data should perform better than random on unseen test data is because the test data is expected to be drawn from the same distribution as the training data. Any given pattern, which is what a subspace represents, may not be applicable in the test set, but a pattern found in multiple disjoint subsets is most likely to be real, and the probability can be mathematically described via the central limit theorem. When applied on an ensemble of models trained with heavily overlapping training sets, the above analysis does not hold, and the ensemble does not perform too much better than the best model. On the other hand, the above algorithm provides a framework from which useful features can be extracted, and this allows larger models to differentiate between true patterns and overfitting.

The analysis above also serves as a justification of most regularization methods as well as an explanation for Occam's Razor in machine learning. "Simple" models are not preferred because they have a higher prior probability than a complicated one--there is no justification for this. "Simple" models are preferred to complicated ones because, in general, they tend to discover regions which are larger and not particularly convoluted and therefore more easily discovered by other models. This, in turn, implies a better bound on the test error. To see why regularization works, it's important to understand what regularization is doing. Regularization effectively forces models to find more similar solutions; for instance, by placing restrictions on weights via the ridge \cite{Hoerl2000RidgeRB} and the lasso \cite{Tibshirani1994RegressionSA}. Thus, models, even when trained on different training sets or with different initializations, are more likely to find similar patterns, and this in turn implies that the patterns which are found are more likely to generalize.

Unfortunately, using regularization means imposing a bias on the model, as per the bias-variance trade-off. These trade-offs are often only backed by intuition and better test error but not justified mathematically. On the other hand, consider an algorithm which follows the general idea as presented above: it trains an ensemble of models; analyzes cluster paths; defines a set of "good" points; and handles "bad" points in some manner, whether by oversampling and training another model or by using some different method. Patterns which were found in smaller models are likely to be discovered by larger models, especially if they are true patterns and are not symptoms of overfitting. These patterns are identified in the larger models, so another model can be trained to focus on the truly bad (the "bad 2") points. This is beneficial for three reasons. First, this ensemble is effectively a form of regularization on the larger models, since it forces the real patterns to be kept, and this increases test accuracy. There is no added bias since no new assumptions were made. Second, this method can be used iteratively, creating a process to continuously increase test accuracy. Third, this provides a framework for how models can collaborate with each other--they should yield when they are unsure and speak up when they are relatively certain. Combined with feature extraction techniques as outlined above, this could allow for an entirely new field of machine learning: continuous learning with neural networks.

\subsection{Bounds on Validation Error of Ensembles}
Consider an ensemble consisting of $k$ neural networks, of which all incorrectly classify at most $v$ of the "good" validation points. Then the number of incorrectly classified validation points when applied only to points such that at least a fraction $f_1$ of the models deem it "good" and of those, at least a fraction $f_2$ agree, is at most $\frac{v}{f_1*f_2}$. This establishes a lower bound on the validation accuracy of the ensemble for "good" points. This is necessary because ensembling effectively trades a decrease in validation accuracy for expanding the number of "good" data points.

\section{Discussion}

\subsection{Oversampling}
The idea behind the model 2’s is this: the model performs significantly better on "good" data points than "bad" data points because "good" data points represent features that the neural network is confident were not created as a result of overfitting. It is natural to continue by creating a new model to focus on the "bad" data points to improve test accuracy. Even if they were classified correctly in the training set, they may still have comparatively high loss. Thus, for any model used to classify "bad" data points, the idea is to increase the weight the model puts on these points in the training set to reduce this loss. Oversampling achieves this effect. On another note, it may be possible perform this process recursively; that is, to continue process for the models in the set of model 2’s and create a set of so-called model 3's.

\subsection{Partitions}
For the training algorithm, the idea of partitioning the training data into a training and validation sets in multiple ways is crucial. This is because the multiple partitions are what create the ability to differentiate between real features and overfitting. Larger models seem to be able to emulate the effects of ensembles of smaller models trained on different portions of the training data. Intuitively, this makes sense--larger models are effectively just smaller models combined together, since neural networks are an iterative model, by design. This is further supported by the fact that for state-of-the-art models, "bad 1" test points are classified with similar accuracy as "original good" test points. Indeed, the critical step occurs in identifying "bad 2" data points, which are outliers in every model, and dealing with these specifically. While the algorithm presented oversamples the corresponding training data, this is not necessarily the optimal approach, and a better method of creating ensembles of models to handle these points should be considered in the future. Finally, it should be noted that the partitions should overlap as little as possible with each other, thus decreasing the probability that the same overfit features would be found by two different models.

\subsection{Justification for Parameters Filtering Clusters}
There exists an intuitive explanation behind the three parameters which were chosen when determining which data points are "good." The distance to a cluster center is considered because all data must belong to a cluster, and therefore clusters contain data which do not really belong to any single cluster. The idea is to filter out these outliers. The number of points in a split is considered because if this is too small, it’s impossible to reason about the accuracy of points which follow that path. Intuitively, this is because points in small splits are most likely outliers which do not truly belong, and while they may be classified correctly in the training set, this is most likely due to overfitting. Finally, the accuracy of a split is taken into consideration because if the accuracy of the split is low in the training or the validation set, there’s no reason to believe it will be better in the test set. 

\section{Conclusion}
In conclusion, this paper presents an algorithm to analyze features of neural networks and differentiate between useful features and those found due to overfitting. This process is then applied to larger models, generating state-of-the-art results for Wide-ResNets on CIFAR-10. Lastly, an analysis concerning bounds for the test accuracy of these ensembles is detailed, and a theorem bounding the test accuracy of neural networks is presented. Finally, the implications of this theorem, including an intuitive understanding of regularization, are presented.

\section{Acknowledgments}
I like to call this project the Miracle Project because it's a true miracle that this project happened...there are too many people to thank. First, thanks to my mom, dad, and brother. Second, thanks to Professors Barnabás Póczos and Majd Sakr. Third, thanks to Theo Yannekis, Michael Tai, Max Mirho, Josh Li, Zach Pan, Sheel Kundu, Shan Wang, Luis René Estrella, Eric Mi, Arthur Micha, Rich Zhu, Carter Shaojie Zhang, Eric Hong, Kevin Dougherty, Catherine Zhang, Marisa DelSignore, Elaine Xu, David Skrovanek, Anrey Peng, Bobby Upton, Angelia Wang, and Frank Li. You guys are the real heroes of this project. And, lastly, thanks to you, my dear reader.
%
%
%
\bibliographystyle{splncs04}
\bibliography{bibtex}

\end{document}